# REAL TIME IMPLEMENTATION OF SPATIAL FILTERING ON FPGA

Chaitannya Supe

Department of Instrumentation & Control Engineering, College of Engineering Pune, Maharashtra, India

## ABSTRACT

*Field Programmable Gate Array (FPGA) technology has gained vital importance mainly because of its parallel processing hardware which makes it ideal for image and video processing. In this paper, a step by step approach to apply a linear spatial filter on real time video frame sent by Omnivision OV7670 camera using Zynq Evaluation and Development board based on Xilinx XC7Z020 has been discussed. Face detection application was chosen to explain above procedure. This procedure is applicable to most of the complex image processing algorithms which needs to be implemented using FPGA.*

## 1. INTRODUCTION

Now a days image processing has become very powerful tool in the field of medical imaging, digital photography, video surveillance etc. Image processing usually requires very large number of operations and high speed data transfer therefore parallel processing or multiprocessing hardware are essential [1]. Because of this, FPGA is one of the best alternatives for image processing as it performs the operations in parallel fashion.

### 1.1. Field programmable gate array

Field Programmable Gate Array (FPGA) is basically pre-fabricated digital logic chips which can be programmed as many times as we want to get desired logic function. FPGAs basically consists of logic cells, interconnects and I/O, all of which can be programmed by user. Logic cell is a basic element of FPGA which can implement any logic function. It consists of look up table, D-flip flop and multiplexer. Interconnect acts as a wire in between logic cells and I/O so as to implement complex logic function. In addition, there are dedicated resources like block RAMs, FIFOs provided by FPGA manufacturer which are used to perform specific task.

### 1.2. Spatial filtering

A spatial filtering on image is neighbourhood operation which changes the value of any pixel by a predefined function of the values of pixels in a neighbourhood of that pixel. For example consider a pixel 'x' having 9 neighbourhood pixels, when spatial filtering is applied, the value of pixel 'x' will be calculated according to the filter function defined for its 9 neighbourhood pixels. Spatial filtering can be used to perform some of the significant image operations such as edge enhancement, image sharpening and noise reduction. These operations can be linear or nonlinear. If the operation performed on the pixels are linear then the corresponding filter is called a linear filter otherwise it is a nonlinear filter [2]. In this paper, implementation of only linear spatial filter has been discussed.

## 2. PROBLEM STATEMENT

In this paper, a step by step approach to apply the spatial filter on FPGA over a real time video frame using a moving window operator is explained. The procedure has been explained by considering face detection as an application. Zynq evaluation and development board (Zed board) has been used as a platform for implementation. Moving window operator of 7x7 rectangular size has been considered due to limitation of memory on Zed board.

## 2. HARDWARE IMPLEMENTATION

In this paper, as mentioned above Zynq Evaluation and Development board commonly known as Zedboard was chosen for implementation of spatial filtering. The ZedBoard is an evaluation and development board based on the Xilinx XC7Z020-1CLG484C Zynq-7000 All Programmable SoC (AP SoC). It has a dual Corex-A9 Processing System (PS) with 85,000 Series-7 Programmable Logic (PL) cells.

The FPGA image processing setup considered in this paper consisted of OV7670 camera from Omni-vision, Xilinx JTAG connector, Zed-board, Zed-board power supply, DB-15 connector and connecting wires as shown in figure 1. There are two JTAGs available on zed board. Here the Xilinx platform cable II has been used which connects Zedboard to PC. Camera was connected to the JA1 and JB1 ports of Zedboard using jumper wires and single stranded wires. VGA output port of Zedboard was connected to monitor through DB-15 connector.

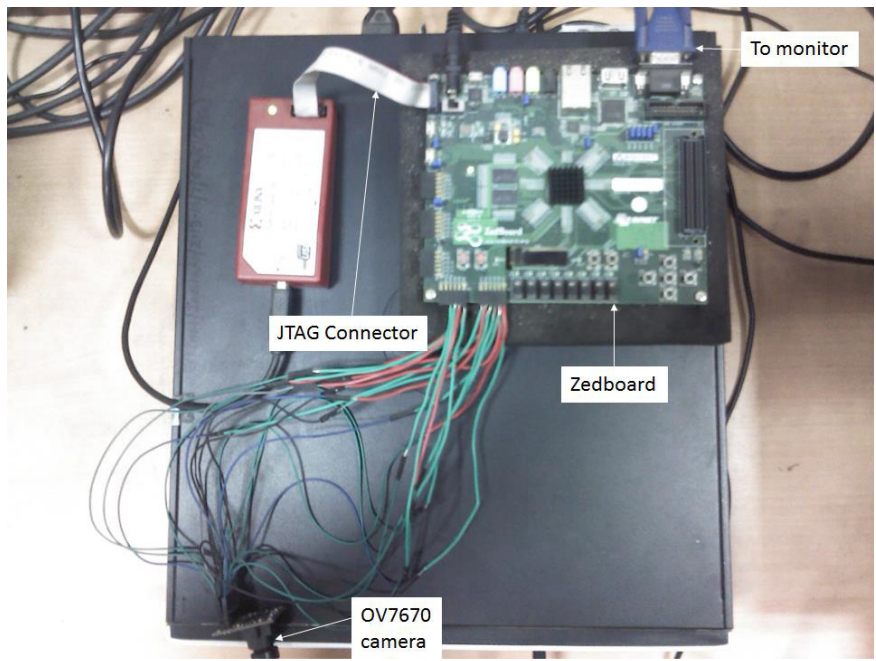

Figure 1.  FPGA image processing setup

The OV7670 camera used here is a low cost, low voltage CMOS image sensor with 18 (9x2) pin package which operates at maximum of 30 fps. It has highest resolution of 640 x 480 (VGA) which is equivalent to 0.3 megapixels. It has built-in digital signal processor which processes the image before sending to the Zedboard. The pre-processing of image is done by digital signal processor via serial camera control bus (SCCB) by setting different values for device control registers which are contained by OV7670 camera module. Camera I/O pins with each pin's description is shown in table 1. The VHDL code for camera interfacing with zed board was taken from [3] and some of the bugs in the code were modified. The coding was done in such a way that outputs of camera module were inputs to the Zedboard and outputs of Zedboard were inputs to the camera. As this paper primarily focuses on spatial filtering, the camera interfacing part was skipped. The detailed information for camera interfacing can be found on [3]. As Zedboard has 12-bit colour output (4 bits each for red, green and blue colour) RGB444 output format of OV7670 was chosen by setting appropriate registers.

Table 1. OV7670 Pins with description.

| Pin | Type | Description |
| --- | --- | --- |
| 3V3, GND | Power | 3.3 volts power supply |
| SIOC | Input | SCCB serial interface clock input |
| SIOD | Input | SCCB serial interface data I/O |
| XCLK | Input | Camera clock input |
| PWDN | Input | Power down mode selection |
| RESET | Input | Resets all registers to default value |
| D[7:0] | Output | YUV/RGB video component output |
| HREF | Output | Horizontal synchronization |
| VSYNC | Output | Vertical synchronization |
| PCLK | Output | Pixel clock output |

To start with actual filtering process with respect to face detection application, first of all, 12-bit camera output data was converted from RGB444 to YUV444 colour space which was accomplished with the help of reversible component transformation (RCT) equations from [4]. Converting the skin pixel information to the modified YUV colour space is more advantageous since human skin tones fall within a certain range of chrominance values (i.e. U-V component) regardless of the skin type.

### 3.1. Thresholding

After converting the video frame into YUV colour space, threshold values were applied only on U component for skin pixel segmentation because of two reasons. First, for skin colour, contribution of red colour is more than green and blue colour contribution is least amongst all [5]. Secondly, Y component is associated only with brightness and not colour therefore only U component was sufficient for skin pixel segmentation. Threshold of $10 < U < 74$ was applied on video frame by converting range from standard 8-bit values to 4-bit values because Zedboard supports 4-bit colour.

In figure 2, the RTL schematic of actual thresholding module written in VHDL is shown. As shown it has two inputs and one output. Din (11:0) represents incoming 12 bit data from camera with respect to positive edge of 25MHz clock (clk). Thresholding operation was applied on this data by using if-else construct of VHDL and output (0:0) was then assigned a single bit value. Here output (0:0) is single bit vector which was explicitly defined because it was given as a input to block RAM ipCore namely frame_buffer in figure 2, which supports input and output data in form of vectors only even for a single bit.

After thresholding, the resulting data needed to be stored hence it was stored in simple dual port block RAM (frame_buffer). Here output (0:0) is written to the block RAM according to write address i.e. valid pixel's address from camera and then the stored data was read by spatial filter unit according to read address i.e. address for displaying the pixels on VGA monitor. Here write address and read address corresponds to "wraddress" and "rdaddress" respectively in figure 2. Both the addresses are 19 bit vectors to address 640x480 = 307200 number of pixels.

## 3.2. Spatial filtering

This step was similar to the erosion operation used in MATLAB which basically shrinks the object. In this, for every pixel p, its neighbouring pixels in a 7x7 neighbourhood were checked. If more than 75% i.e. 37 of its neighbours were skin pixels, then p was declared a skin pixel. Otherwise p was declared a non-skin pixel. This stage allowed most background noise to be removed.

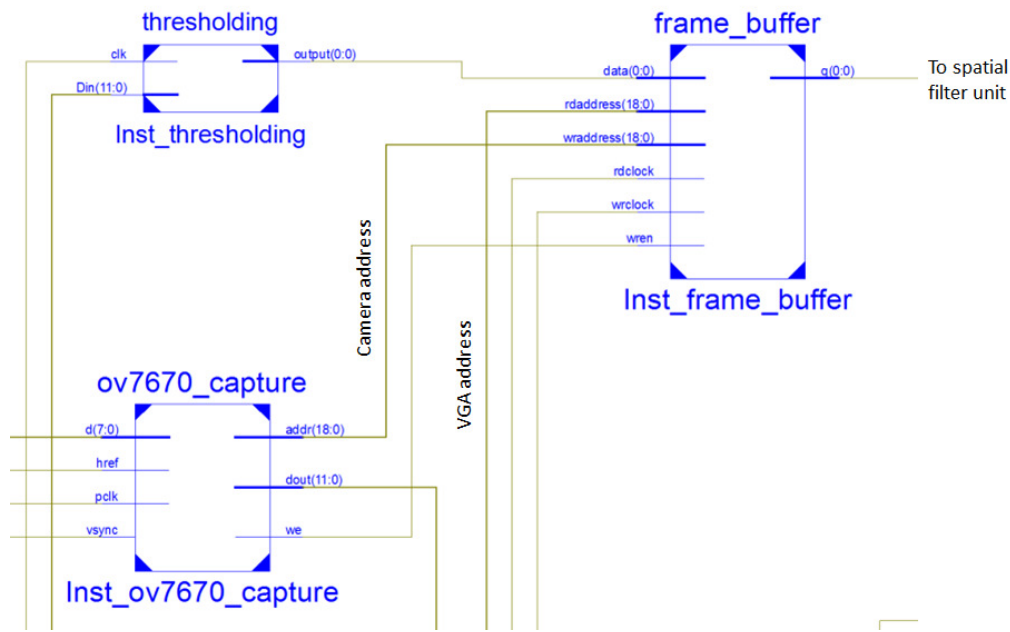

Figure 2. Thresholding VHDL module with connections

Now to implement this module in Zedboard, the module was subdivided into two modules. First module to generate 7x7 window and second to calculate the pixel's value from window as shown in figure 3. The 7x7 window was not possible to show in the figure hence generic NxN window is shown. Each tiny square represent a pixel's location with respect to window. Row buffers shown in the figure 3 are FIFO memories created using Xilinx FIFO generator core. NxN window correspond to 7x7 window in this case.

### 3.2.1. Window generation

To examine the neighbours around a pixel, their values needed to be stored. Therefore 6 FIFOs were created to buffer 7 consecutive rows of each video frame. In general for NxN window N-1 numbers of FIFOs should be created. The structural view of Xilinx FIFO core is shown in figure 4. The din (0:0) and dout (0:0) are single bit input and output vectors respectively. The wr_en and rd_en are write and read enable pins respectively. The data_count (9:0) represents 10-bit data count to count number of pixels stored inside the FIFO. Common clock configuration of FIFO was considered for synchronous operations.

FIFO and registers were used to provide an internal line delay and to synchronize the supply of input pixel values to the processing elements i.e. to window operator module as shown in figure 5, ensuring that all the pixel values involved in a particular neighbourhood operation were processed at the same instant thus achieving a parallel pipelined structure [6, 7]. All 'rxx' elements in figure 5 represent registers used for pipelining the data coming from frame_buffer module. The 'wxy' represents the pixel values in xth row and yth column of the window.

The binary output data from frame_buffer module was sent to series of registers and FIFOs as per figure 5. The length of each row of video frame was 640 pixels out of which 633 pixels were stored in FIFO memory and remaining 7 were stored in registers, hence Memory depth of 1024 was chosen as Xilinx FIFO core does not have custom 633 memory depth. Now as the process was real-time, a condition was created to store exact 633 pixel values in each FIFO. Whenever data count of a FIFO reaches 631, the read enable signal for that FIFO and write enable signal for next FIFO were set to '1'. The value 631 was chosen instead of 633 to compensate with latency of 2 clock cycles generated.

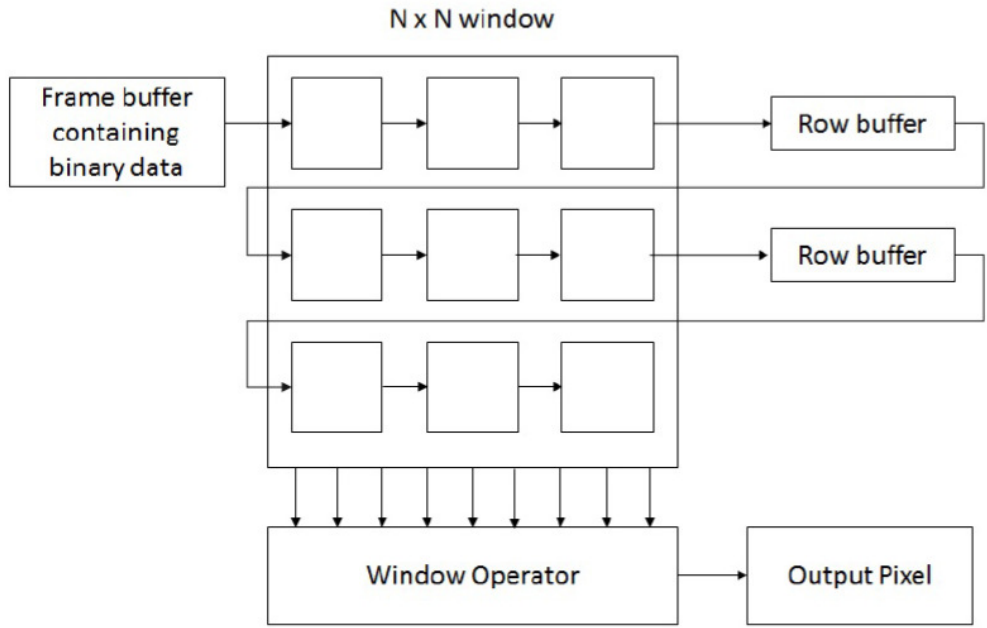

Figure 3. Architectural of spatial filter module

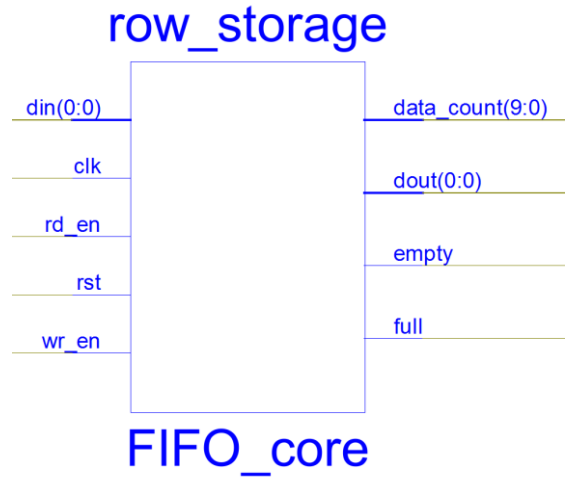

Figure 4. Xilinx FIFO ipCore structural view

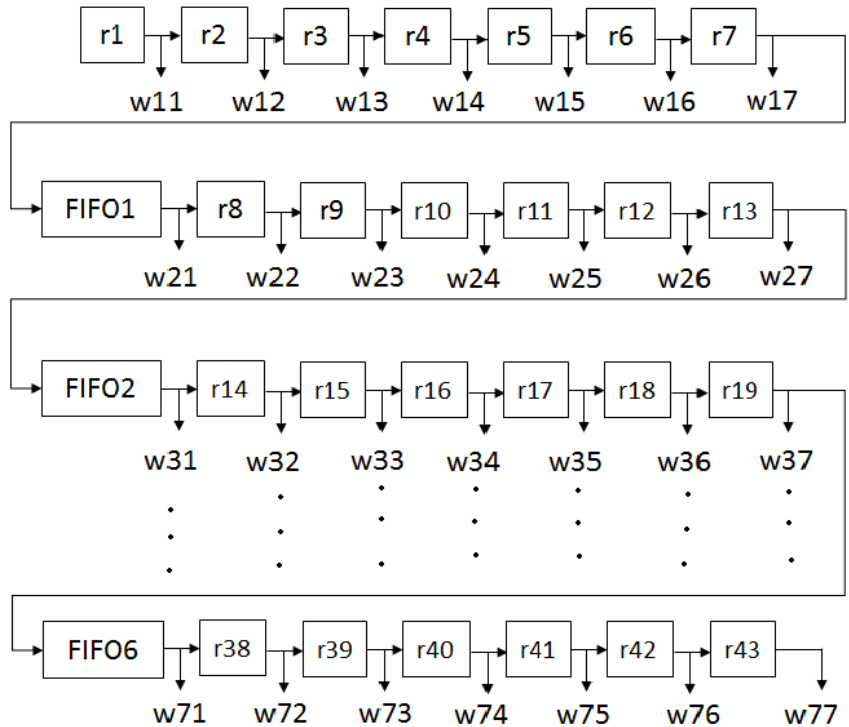

Figure 5. Architecture of window generator

### 3.2.2. Window operator

The window operator module processes one pixel at a time i.e. per clock cycle, changing its value by a specified function in the 7x7 window around the pixels [8, 9]. Window operator module operates on stored 7x7 window i.e. the outputs of window generation module to output skin or non-skin pixel. In this, all the pixel values from window generation module i.e. w11 to w77 were simply added and if their sum is greater than 37 then output pixel will be assigned a single bit value of '1' else '0' was assigned. In FPGA, some invalid pixel values get generated while processing on a real time video frame because of clock cycle latencies, to remove these values in window operator module a data valid signal was created in such a way that it will let the data pass only after 48 clock cycles as the window is of 7x7 = 49 pixels. The main reason behind creating such signal is parallel processing of FPGA. The window operator therefore starts processing a pixel only after data valid signal becomes '1'.

## 4. CONCLUSIONS

It is concluded that the proposed design is pipelined so it operates at much higher speed than normal non-pipelined design which is very useful in complex image processing algorithms. Also design uses very less amount of FPGA dedicated resources as the operations were carried out on binary data. Such type of operations are needed in industrial applications such as visual inspection, sensor-controlled handling and assembly, control of tools, machines, also in military applications where detection of target is main criteria. The whole procedure will also be useful for academic purposes. Figure 6 shows device utilization summary for window generation module alone.

| Device Utilization Summary (estimated values) | | | [-] |
|---|---|---|---|
| Logic Utilization | Used | Available | Utilization |
| Number of Slice Registers | 458 | 106400 | 0% |
| Number of Slice LUTs | 390 | 53200 | 0% |
| Number of fully used LUT-FF pairs | 198 | 650 | 30% |
| Number of bonded IOBs | 53 | 200 | 26% |
| Number of Block RAM/FIFO | 3 | 140 | 2% |
| Number of BUFG/BUFGCTRLs | 1 | 32 | 3% |

Figure 6. Device utilization summary for window generation module

Figure 7 shows device utilization summary for all modules including camera interfacing modules. It can clearly be seen that number of dedicated resources required for generation of window alone are very less in comparison with all modules.

| Device Utilization Summary (estimated values) | | | [-] |
|---|---|---|---|
| Logic Utilization | Used | Available | Utilization |
| Number of Slice Registers | 617 | 106400 | 0% |
| Number of Slice LUTs | 751 | 53200 | 1% |
| Number of fully used LUT-FF pairs | 369 | 999 | 36% |
| Number of bonded IOBs | 33 | 200 | 16% |
| Number of Block RAM/FIFO | 23 | 140 | 16% |
| Number of BUFG/BUFGCTRLs | 4 | 32 | 12% |

Figure 7. Device utilization summary for project

## ACKNOWLEDGEMENTS

I, Chaitannya Vinay Supe, the post graduate student of College of Engineering, Pune, am extremely grateful to all the people for the confidence bestowed in me and entrusting my work. At this juncture I feel deeply honored in expressing my sincere thanks to Er. Mike Field for his valuable guidance regarding interfacing of camera which was crucial step towards successful completion of my work. I also extend my gratitude to my Project Guide Dr. S. L. Patil, who assisted me in completion of this work. I would also like to thank to Ms. K. A. Bhole, Mr. Vihang Naik, Mr. Divyesh Ginoya for their critical advice and guidance without which this work would not have been possible. Last but not the least I place a deep sense of gratitude to my family members and my friends who have been constant source of inspiration during the preparation of this work.

# Author

**Chaitannya Supe** was born in Malkapur, India, in 1989. He received the B.E. degree in instrumentation and control engineering from the Vishwakarma Institute of Technology, Pune, India, in 2011, and the M.Tech. degree in instrumentation and control engineering from the College of Engineering, Pune, India, in 2014. His current research interests include digital image processing, embedded systems .

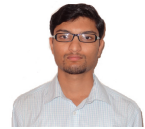